\documentclass[10pt,twocolumn,letterpaper]{article}

\usepackage{cvpr}
\usepackage{times}
\usepackage{epsfig}
\usepackage{graphicx}
\usepackage{amsmath}
\usepackage{amssymb}
\usepackage{makecell}
\usepackage{url}
\usepackage{verbatim}
\usepackage{multirow}
\usepackage{type1ec}
\usepackage{scalerel}
\usepackage{threeparttable}
\usepackage{booktabs, arydshln} 

\usepackage{fancyhdr}

\usepackage[pagebackref=true,breaklinks=true,letterpaper=true,colorlinks=false,bookmarks=false]{hyperref}

\usepackage[acronym]{glossaries}
\newacronym{FPV}{FPV}{First-Person Video}
\newacronym{TPV}{TPV}{Third-Person Video}
\newacronym{YOLO}{YOLO}{YOLO}
\newacronym{DNN}{DNN}{Deep Neural Network}
\newacronym{CNN}{CNN}{Convolution Neural Network}
\newacronym{MIFF}{MIFF}{Multi-Importance Fast-Forward}
\newacronym{GTEA}{GTEA Gaze+}{Georgia Tech Egocentric Activity}
\newacronym{ASTAR}{ASTAR}{A*STAR Ego-Gaze}
\newacronym{MSCOCO}{MSCOCO}{Microsoft Common Objects in Context}
\newacronym{SORT}{SORT}{Simple Online and Realtime Tracking}
\newacronym{SAS}{SAS}{Sparse Adaptive Sampling}
\newacronym{NPD}{NPD}{Normalized Pixel Difference}
\newacronym{fps}{fps}{frames per second}
\newacronym{IMU}{IMU}{Inertial Measurement Unit}
\newacronym{GPS}{GPS}{Global Positioning System}
\newacronym{LSTM}{LSTM}{Long Short-term Memory}
\newacronym{DTW}{DTW}{Dynamic-time-warping}
\newacronym{LLC}{LLC}{Locality-constrained Linear Coding}
\newacronym{NFOV}{NFOV}{Normal field-of-view}
\newacronym{FOV}{FOV}{field-of-view}

\def\cvprPaperID{****} 

\makeatletter
\def\adl@drawiv#1#2#3{%
	\hskip.5\tabcolsep
	\xleaders#3{#2.5\@tempdimb #1{1}#2.5\@tempdimb}%
	#2\z@ plus1fil minus1fil\relax
	\hskip.5\tabcolsep}
\newcommand{\cdashlinelr}[1]{%
	\noalign{\vskip\aboverulesep
		\global\let\@dashdrawstore\adl@draw
		\global\let\adl@draw\adl@drawiv}
	\cdashline{#1}
	\noalign{\global\let\adl@draw\@dashdrawstore
		\vskip\belowrulesep}}
\makeatother
\DeclareMathOperator*{\argmin}{argmin} 
\newcommand{\norm}[2]{\left\lVert#1\right\rVert_{#2}}

\cvprfinalcopy

\ifcvprfinal\fi
\begin{document}

\title{A gaze driven fast-forward method for first-person videos}

\author{Alan~Neves,
		Michel~Silva,
		Mario~Campos, and
		Erickson~R.~Nascimento\\
		Universidade Federal de Minas Gerais (UFMG),
		Belo Horizonte, Brazil \\
{\tt\small \{alan.neves, michelms, mario, erickson\}@dcc.ufmg.br}
}

\maketitle

\thispagestyle{fancy}
\fancyhf{}
\chead{\small{Accepted for presentation at the Sixth International Workshop on Egocentric Perception, Interaction and Computing \\ at the IEEE/CVF Conference on Computer Vision and Pattern Recognition (EPIC@CVPR) 2020}}
\setlength{\headsep}{0.35 in}


\begin{abstract}
	The growing data sharing and life-logging cultures are driving an unprecedented increase in the amount of unedited First-Person Videos.
	In this paper, we address the problem of accessing relevant information in First-Person Videos by creating an accelerated version of the input video and emphasizing the important moments to the recorder. Our method is based on an attention model driven by gaze and visual scene analysis that provides a semantic score of each frame of the input video.
	We performed several experimental evaluations on publicly available First-Person Videos datasets. The results show that our methodology can fast-forward videos emphasizing moments when the recorder visually interact with scene components while not including monotonous clips.
\end{abstract}

\section{Introduction}
\label{sec:introduction}

The advent of wearable devices and their crescent popularization, combined with the unlimited cloud storage capacity, led to massive growth in the available first-person video data over the Internet.
People are spreading their achievements, advertising hazards, recording criminal evidence in law enforcement scenarios, and logging their daily activities --- generating uncountable hours of long and monotonous \glspl{FPV}.

In general, long \glspl{FPV} are composed of relevant and non-relevant events, categorized according to what the person who recorded the video saw and experimented during the recording. 
Recent techniques address the problem of providing quick access to the information through techniques such as summarization~\cite{Xu:2015} or semantic fast-forward~\cite{Silva2018}. These techniques evaluate and prune out less relevant frames, creating a shorter version of the input video. However, they are limited by predefined semantic that bounds the application to specific purposes tasks.

The human vision focus, \ie, Gaze, provides cues of the wearer's overt attention. Such information allows us to infer what is relevant to a person at a specific moment, without being previously determined. 
In this paper, we propose a gaze-based semantic hyperlapse method for \glspl{FPV} that emphasizes video segments by inferring the recorder's attention during the recording.
The recorder's attention is modeled fusing gaze and visual scene components information. Furthermore, we model the novelty of the scene by penalizing long unvarying video segments.

\paragraph{Related Work.}

Summarization is a popular approach to produce a shortened version of an \gls{FPV} video.
These methods define, rank, and select video segments concerning a definition of what is relevant, composing a summarized version of the input video.
In {\it viewer-based} methods~\cite{de2011vsumm,Lu2013}, video features are used to perform the summarization, as oppose to the {\it important to the wearer}~\cite{Aizawa2001,Xu:2015} methods that infer relevance from cues of the camera wearer intentions.

However, video summarization techniques do not take into account temporal continuity and visual smoothness of the summarized video.
Semantic hyperlapse techniques, for their turn, emphasize relevant parts of the video while they can handle these restrictions. Usually, the effect of emphasizing the semantic information is produced by reducing the speed-up rate of the relevant video segment or zooming in on the region of the image containing the relevant content.
Most of the relevance definition applied in semantic Hyperlapse works rely on a set of predefined objects, such as the presence of faces~\cite{Ramos:2016}, pedestrians~\cite{Silva2018cvpr, Silva2018, Silva2020tpami}, street crosswalk~\cite{Okamoto2014}, or use of machine learning to identify user's preferences~\cite{Silva2018}.

Predefined semantic-based techniques are unable to capture the dynamic behavior of the wearer when assigning the frame relevance. To tackle this lack of certainty about the wearer's focus, we propose to use the gaze in the context of the hyperlapse problem.
The gaze is defined as the region of the visual field where the fovea points to, in order to capture detailed information. As the visual field is wide and fovea is small, eye movements must be employed to move the fovea to desired targets, \ie, the focus of the camera wearer.

In this work, we fast-forward \glspl{FPV} by selecting relevant moments to the recorder based on his/her visual interaction with scene components through a novel gaze-based method.

\begin{figure*}[!t]
	\centering
	\includegraphics[width=.9\linewidth]{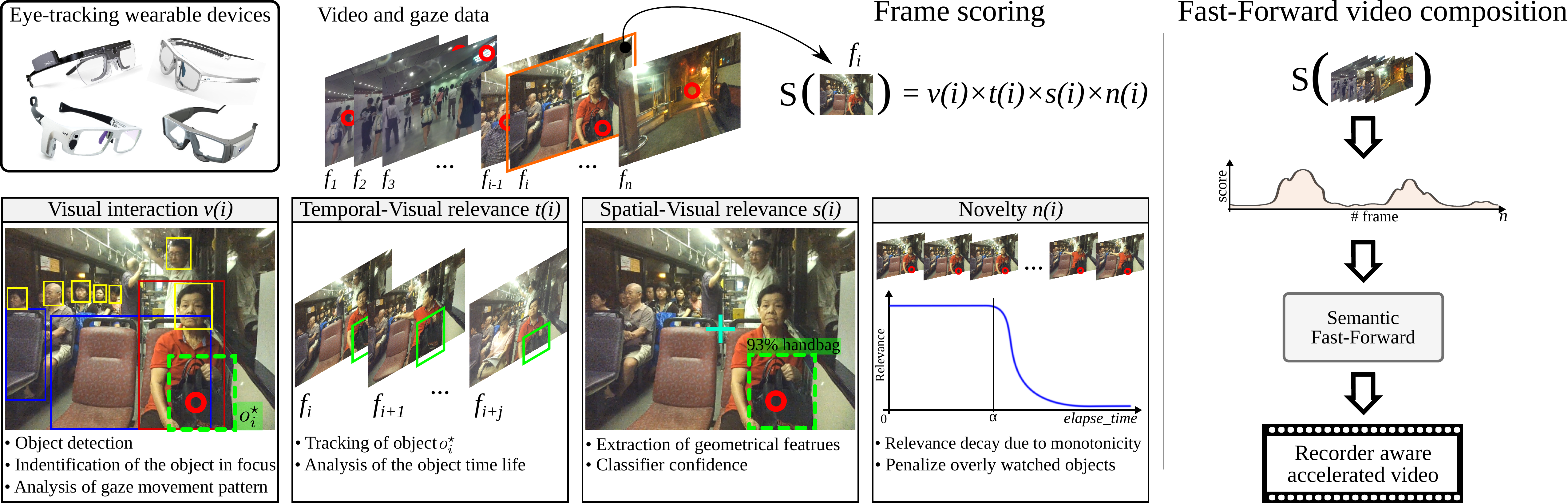}
	\caption{Gaze-based semantic fast-forward methodology. The object in focus $o_i^\star$ is identified. Its trajectory is tracked in the temporal visual relevance step. Visual and geometrical features also contribute to the frame scoring in spatial visual relevance step. A novelty term penalizes objects that have been focused for a long time. After scoring all frames of the video and feeding the adaptive frame selection technique, a new video is created, emphasizing the relevant parts to the wearer.}
	\label{fig:methodology}
\end{figure*}

\section {Methodology}
\label{sec:methodology}

Our methodology is based on four components as depicted in Fig.~\ref{fig:methodology}: $i$)  visual interaction; $ii$) temporal and $iii$) spatial relevance; and $iv$) avoidance of over watching video portions.

\paragraph{Visual interaction.}
\label{paragraph:visual_intercation}

Let $f_i$ be the \textit{i}-th frame of the video, ${\mathbf{g}_i={\left[x,y\right]^T}}$ be the gaze position regarding frame $i$ coordinate system, and $\mathcal{O}_i$ the set of bounding boxes of the detected objects in the frame $i$. To infer which object is observed by the wearer, we check whether the gaze position $\mathbf{g}_i$ is lying inside of the bounding boxes of objects in the set $\mathcal{O}_i$.
The result is the set of objects ${\mathcal{O}_k~\subset~\mathcal{O}_i}$. The smallest area non-occluded object, assumed as the foreground object, is obtained by:

\begin{equation}
o_i^\star = \argmin_{o_{j}~\in~\mathcal{O}_k} A(o_j^{bb}),
\end{equation}

\noindent where the function ${A(x^{bb})}$ returns the area of the bounding box $x^{bb}$ around the object $x$.

The eye-tracking system provides the gaze movement pattern $mp(\mathbf{g}_i)$ of the $ i $-th frame, used to express our attention level in the visual interaction $v(i)$ term: 
\begin{equation} \label{eq:gaze_mov}
v(i) =
\begin{cases}
1.0, & \text{if } mp(\mathbf{g}_i) = \text{Fixation}; \\
0.5, & \text{if } mp(\mathbf{g}_i) = \text{Saccade}; \\
0.0, & \text{if } mp(\mathbf{g}_i) = \text{Blink or}~\mathcal{O}_k = \emptyset. \\
\end{cases}
\end{equation}

\paragraph{Temporal-Visual relevance.}

It is expected that the objects of interest remain longer in the recorder's field of view. We model this behavior tracking objects that visually interact with the user along with the video, from the start of the visual interaction with $o_i^\star$ until attention got dispersed.

The temporal-visual relevance is calculated as 

\begin{equation}
t(i) = \frac{tr(o_i^\star, \text{duration})}{T_{max}},
\end{equation}

\noindent where ${tr(o,\text{duration})}$ indicates the tracking duration of the object $o$, and $T_{max}$ is the longest tracking duration considering all objects of a video $V$ composed of $n$ frames.

\paragraph{Spatial-Visual relevance.}

We embedded geometrical and visual features to describe image importance based on gaze and object detector information, as follows:

\begin{itemize}
	\item \textit{Relative area:} Ratio of the focused object $o_i^\star$ bounding box area \wrt the whole image: ${a_{o_i^\star} = {A({o_i^\star}^{bb})}/{A(f_i)}}$. ${x^{bb}}$ indicates the box around object $x$,  ${A(.)}$ is the bounding box area function and $A(f_i)$ is the relative area of the \textit{i-th} frame $f_i$. \item \textit{Centrality:} 
	Egocentric gaze has a strong center bias related to many aspects involving human interaction with the world.
	We model it as defined:
	\begin{equation}
	c_{o_i^\star} = \left(1 + \norm{C({o_i^\star}^{bb}) - C(f_i)}{2}\right)^{-1},
	\end{equation}
	\noindent where ${C(.)}$ returns the central point of a rectangle.
	
	\item \textit{Focus:} The horizontal distance between focused object $o_i^\star$ center and gaze coordinates $\mathbf{g}_i$, given by:
	\begin{equation}
	m_{o_i^\star} = \left(1 + \norm{\mathbf{g}_i - C({o_i^\star}^{bb})}{h}\right)^{-1},  
	\end{equation}
	\noindent where $ \norm{.}{h}$ is the $L_{2}\ \text{norm}$ regarding the x-axis. 
	
\end{itemize}

The final spatial-visual relevance score $s_i$ of the $i$-th frame is given by the sum of the three previous defined geometric features weighted by the classifier confidence~($ d_{o_i^\star} $):
\begin{equation}
s(i) = d_{o_i^\star}(a_{o_i^\star} + c_{o_i^\star} + m_{o_i^\star}). 
\end{equation}

\paragraph{Novelty model.}
\label{subsec:novelty_model}

When performing long tasks that require visual focus, \eg, watching TV, even with the wearer's visual interaction, the task itself might not be relevant in its entirety. To avoid overly watching these segments, we propose a weighting factor $n(i)$ based on the tracking of object $o_i^\star$ through a conditional exponential decay function:
\begin{equation} \label{eq:novelty}
n(i) =
\begin{cases}
1 & , \text{if } et_i < \alpha; \\
e^{\scaleto{-\frac{et_i~-~\alpha}{2}}{14pt}} & , \text{otherwise}, \\
\end{cases}
\end{equation}

\noindent where $et_i = i - tr(o_i^\star, \text{start time stamp})$ and it refers to the focus elapsed time, while $tr(o_i^\star, \text{start time stamp})$ returns the initial frame when the user first focused in object $o_i^\star$. The condition $\alpha$ models when to consider new information.

\paragraph{Adaptive frame selection.}

In the last step, we score each frame $i$ according to 

\begin{equation}
\label{eq:frame_score}
S_i = v(i)~t(i)~s(i)~n(i),
\end{equation} 
\noindent combining the previous four scores through product operation, ensuring dependency among them. Then, we feed the state-of-the-art techniques regarding semantic adaptive frame selection with our gaze-based semantic profile $S$.

The adaptive frame selection works by segmenting the video into relevant and non-relevant segments and calculating different speed-up rates for them, emphasizing the most relevant by assigning lower rates; regarding smoothness, semantic information, and video size. The reader is referred to~\cite{Silva2018,Silva2018cvpr} for more details on the adaptive frame selection.

\section {Experiments}

\paragraph{Datasets.}
We evaluated our approach on the \gls{GTEA} and \gls{ASTAR} datasets.
The \gls{GTEA} dataset\footnote{Publicity available at \href{http://www.cbi.gatech.edu/fpv}{www.cbi.gatech.edu/fpv}}~\cite{Fathi:2012} comprises videos of seven meal preparation activities performed by different subjects. It contains gaze data, actions, and hand masks annotations.
The \gls{ASTAR}~\cite{Ma:2012}\footnote{Available under request.} is an unconstrained egocentric dataset, recorded in an opened scenario, where the participants perform free daily activities. It contains gaze data annotations and overall information about actions performed throughout the video.

\paragraph{Competitors Methods.}
\label{sec_baseline}
We compared our method against two state-of-the-art semantic hyperlapse techniques: the \gls{MIFF}~\cite{Silva2018} and \gls{SAS}~\cite{Silva2018cvpr}.
Due to the absence of faces on videos of the used datasets, we changed the default semantic definition in the \gls{MIFF} method from face to object presence. After detecting the objects using the \acrshort{YOLO} network, frame scoring is performed as proposed by the authors of the original paper.

\paragraph{Results and Discussion.}
To evaluate the quality of the frame sampling, motivated by the work of Xu~\etal~\cite{Xu:2015}, we measured the overlap between the emphasized segments and the ground truth task segments that demands the user's visual interaction, the high attention tasks.
To select these tasks, we performed a filtering step on ground truth tasks, selecting only those on which the eye-tracking monitor logged at least $50\%$ of the task duration as gaze fixation. 
We consider an emphasis whenever the acceleration rate applied to the segment is smaller than half of the video required speed-up.
The \textit{emphasized actions} metric counts the overlaps between emphasized segments on the accelerated video and high attention tasks, given by $Ea = {\left | E \cap \mathcal{H} \right |}$,
where $\mathcal{H}$ is the set of high attention tasks of the video, $E$ is the set of true emphasized segments on accelerated video and $\left | . \right |$ is the cardinality of a set. 
A large overlapping rate indicates that the user visually interacted with scene components.

We report the results for the \textit{emphasized actions} values as the mean over the videos in \gls{GTEA} dataset. The accelerated videos were generated using \gls{MIFF} semantic hyperlapse technique with three different semantic definitions: \textit{CoolNet}~\cite{Silva2018}, YOLO, and Ours.
\acrshort{YOLO} semantics presented the worst result, emphasizing only $7.2\%$ of the high attention tasks, since it detects many kitchen elements on the scene without emphasizing any specific object at all, resulting in a fast-forwarded video with high speed-up. 
\textit{CoolNet} emphasized $7.6\%$ of the high attention tasks while being unable to capture and emphasize user attention since it is focused on identifying visual appealing frames.
Our method, thanks to the modeling of the recorder visual interaction, achieved the best result emphasizing $17.2\%$ of the high attention tasks.

We also evaluated the quality factors of the produced video using the \textit{Instability} and \textit{Speed-up} metric proposed in the work of Silva~\etal~\cite{Silva2018}.
Tab.~\ref{tab:fast_forward_analysis} presents the mean values over all videos for both datasets. The accelerated videos were generated using \gls{SAS} and \gls{MIFF} techniques combined with the three semantic definitions: \textit{CoolNet}, \acrshort{YOLO}, Ours. The results demonstrate that our gaze-based frame scoring is feasible to be combined with the semantic fast-forward techniques since it did not negatively affect their results.

\begin{table}[t]
	\setlength{\tabcolsep}{1.50pt}
	\small
	
	\centering
	\caption{Results for \textit{Instability} and \textit{Speed-up} metrics using different score techniques.}
	\label{tab:fast_forward_analysis}
	\begin{threeparttable}[t]
		\begin{tabular}{clccccccc}
			\toprule
			& \multirow{2}{*}{\textbf{Dataset}} & \multicolumn{3}{c}{\textbf{\gls{SAS}}} & & \multicolumn{3}{c}{\textbf{\gls{MIFF}}} \\
			& & \textit{CoolNet}	& \acrshort{YOLO} &	Ours			& & \textit{CoolNet}	&	\acrshort{YOLO}			&	Ours	\\
			\cline{3-5} \cline{7-9} 
			\multirow{3}{*}[-0.5em]{\rotatebox[origin=c]{90}{\textit{Instability}\tnote{1}}} \\
			& \gls{ASTAR}			& $30.5$			& $32.5$			& $\mathbf{30.3}$	& & $33.7$		& $35.3$				& $32.5$	\\
			& \gls{GTEA}			& $22.3$			& $\mathbf{20.7}$	& $23.2$			& & $27.2$		& $27.4$				& $28.7$	\\
			& \multicolumn{8}{l}{\scriptsize{$^1$Lower is better.}}
			\\
			\cdashlinelr{1-9}
			\multirow{3}{*}[-0.5em]{\rotatebox[origin=c]{90}{\textit{Speed-up}\tnote{2}}} \\
			& \gls{ASTAR}			& $\mathbf{0.1}$ 	& $\mathbf{0.1}$	& $0.2$			& & $0.3$		& $1.6$				& $0.4$	\\
			& \gls{GTEA}			& $0.5$			& $0.8$			& $0.3$			& & $0.2$		& $\mathbf{0.0}$		& $0.1$	\\
			& \multicolumn{8}{l}{\scriptsize{$^2$Better close to zero.}}
			\\
			\bottomrule
		\end{tabular}
	\end{threeparttable}
\end{table}

\begin{figure}[t]
	\centering
	\includegraphics[width=\linewidth]{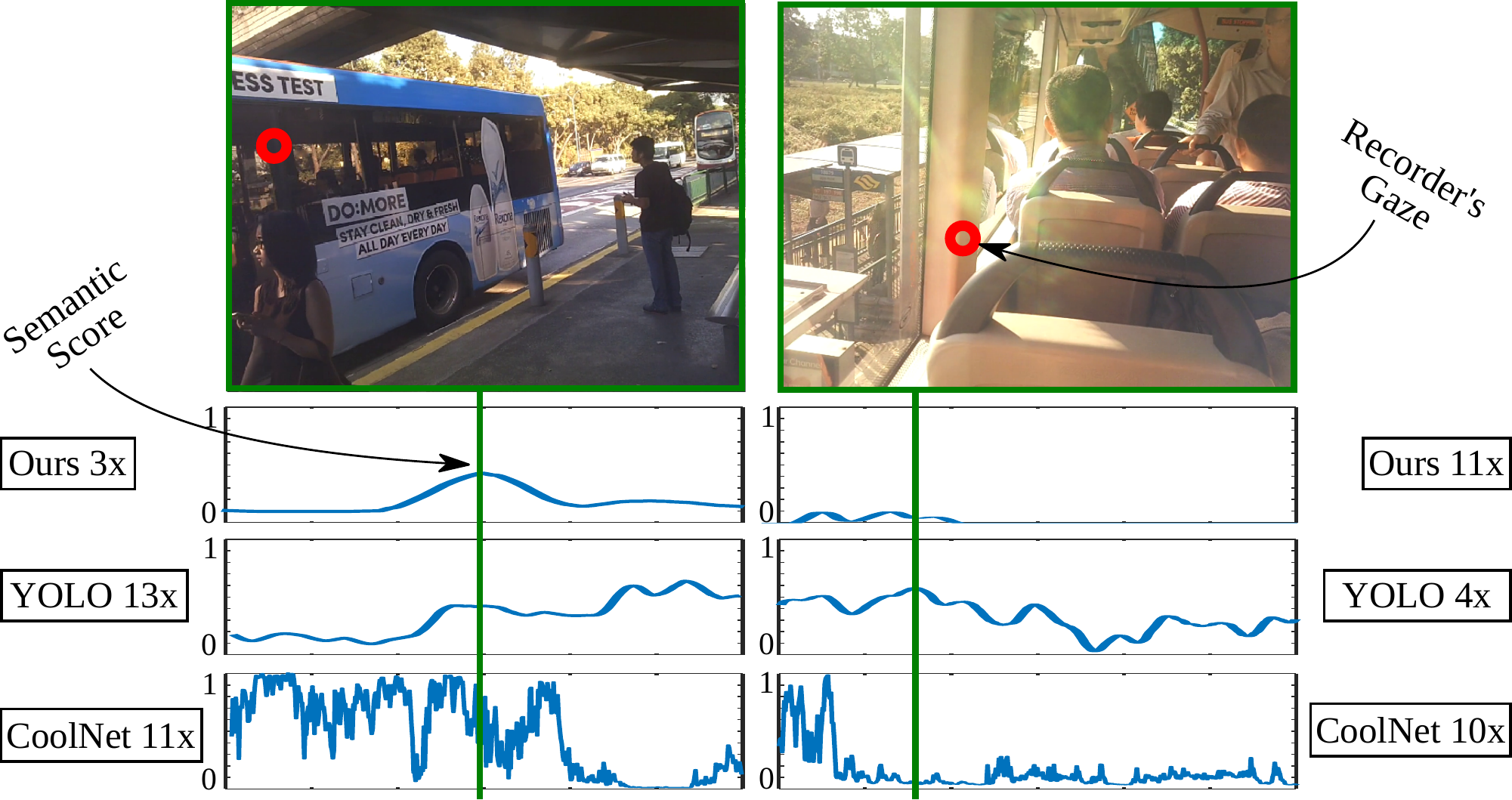}
	\caption{Example of two scenes and the semantic score computed by three different semantic information extraction approaches. Wearer's gaze is presented as a red circle.}
	\label{fig:results_fig2}		
\end{figure}

Qualitative results are presented in Fig.~\ref{fig:results_fig2}, with the semantic values and the speed-up for two specific frames from video \textit{S003\_C02} of the \gls{ASTAR} dataset. The frame on the left received a high value in our model due to the visual interaction between the recorder and the approaching bus. A crescent value is assigned when using \acrshort{YOLO}-based semantic definition due to the bus detection and its crescent proximity throughout the time. \textit{CoolNet} outputs high semantic values due to the presence of trees in the background.
In the frame on the right, our method assigned low values of relevance since there is no visual interaction between the user and scene components.
The \acrshort{YOLO}-based semantics returned a higher score due to the presence of several objects on the scene. For CoolNet, the low scores obtained should be related to the indoor-like scenes, as the interior of the bus.
\acrshort{YOLO} detected many objects in both images, leading to high speed-up rates on interesting segments. 
CoolNet returned large scores to many regions of the video, being unable to emphasize specific moments of them.
Both images contain distinguishable scene elements, but our method remains unaffected by their presence unless they arouse the interest of the wearer.

\section{Conclusion}

We presented a new attention model based on the fusion of gaze and visual information, and a scene novelty modeling used for emphasizing relevant moments to the camera wearer on \glspl{FPV}.
We evaluated our approach in two datasets, against two semantic hyperlapse methods.
Our method showed a better average result of $9.6$ p.p. to the best competitor in the coverage of high attention tasks. 
Also, our method does not negatively impair the frame sampling step from the used hyperlapse methods.
Visual results presented our method ability to capture the user's intentions properly.
One limitation of the proposed method is the requirement of egocentric gaze data, which we are addressing in order to make the methodology applicable to the general video.

{\small
\bibliographystyle{ieee_fullname}
\bibliography{refs}
}

\end{document}